\documentclass[letterpaper, 10 pt, conference]{ieeeconf}  

\IEEEoverridecommandlockouts                            

\usepackage{graphics} % for pdf, bitmapped graphics files
\usepackage{epsfig} % for postscript graphics files
\usepackage{mathptmx} % assumes new font selection scheme installed
\usepackage{times} % assumes new font selection scheme installed
\usepackage{amsmath} % assumes amsmath package installed
\usepackage{amssymb}  % assumes amsmath package installed
\usepackage{outlines}
\usepackage{xcolor}
\usepackage[ruled,vlined,linesnumbered]{algorithm2e}
\usepackage{gensymb}
\usepackage{caption}
\usepackage{float}
\usepackage{subcaption}
\usepackage[colorlinks=true,linkcolor=black,anchorcolor=black,citecolor=black,filecolor=black,menucolor=black,runcolor=black,urlcolor=black]{hyperref}
\usepackage{siunitx}
\usepackage{newtxmath}  % use original mathcal
\usepackage{mathtools}
\usepackage{cite}
\usepackage{algpseudocode}

\title{\LARGE \bf
Multi-Robot Collaborative Localization and Planning with Inter-Ranging
}

\author{Derek Knowles$^{1}$, Adam Dai$^{2}$, and Grace Gao$^{3}$% <-this % stops a space
\thanks{$^{1}$Department of Mechanical Engineering,
        Stanford, CA 94305, USA,
        dcknowles@stanford.edu}%
\thanks{$^{3}$Department of Electrical Engineering,
        Stanford, CA 94305, USA,
        addai@stanford.edu}%
\thanks{$^{4}$Department of Aeronautics and Astronautics,
        Stanford, CA 94305, USA
        gracegao@stanford.edu}%
}

\newcommand{\norm}[1]{\left\Vert#1\right\Vert}
\newcommand{\invsub}[1]{\left( #1 \right)^{-1}}

\begin{document}

\maketitle
\thispagestyle{empty}
\pagestyle{empty}

\begin{abstract}

Robots often use feature-based image tracking to identify their position in their surrounding environment; however, feature-based image tracking is prone to errors in low-textured and poorly lit environments.
Specifically, we investigate a scenario where robots are tasked with exploring the surface of the Moon and are required to have an accurate estimate of their position to be able to correctly geotag scientific measurements.
To reduce localization error, we complement traditional feature-based image tracking with ultra-wideband (UWB) distance measurements between the robots.
The robots use an advanced mesh-ranging protocol that allows them to continuously share distance measurements amongst each other rather than relying on the common ``anchor" and ``tag" UWB architecture.
We develop a decentralized multi-robot coordination algorithm that actively plans paths based on measurement line-of-sight vectors amongst all robots to minimize collective localization error.
We then demonstrate the emergent behavior of the proposed multi-robot coordination algorithm both in simulation and hardware to lower a geometry-based uncertainty metric and reduce localization error.

\end{abstract}

\section{Introduction}
\label{jpl:intro}

Robots for extraplanetary surface exploration must have precise localization capabilities to perform productive science, accurately geotag scientific measurements, and navigate safely.
Currently, robots that explore extraplanetary surfaces often use cameras and feature-based tracking called visual odometry to localize themselves~\cite{johnson_mars_lander, rankin_mobility}.
However, visual odometry is prone to errors when there are few features to track due to poor lighting or low-textured regions~\cite{8764393}.
Hence, it is desirable to provide alternate or complementary means to deal with measurement errors from camera-based systems.

If rather than a single robot exploring, there is a team of collaborative robots exploring an area, then one additional method for reducing localization uncertainty is to share helpful information among the robots~\cite{7148524, yan2013survey, 1677943}.
Specifically, robots may measure distances between themselves and neighboring robots, a technique which we refer to as inter-ranging~\cite{qiang_formation_uwb,stier2022}.
These distance measurements, combined with the positions of neighboring robots, may then be used to triangulate position and reduce localization uncertainty.

However, the benefit of performing inter-ranging localization is greatly dependent on the geometry of the neighboring robots.
Several metrics exist that quantify the information gain available from a set of distance measurements including the Cramer-Rao lower bound~\cite{stoica2021cramer} 
and the Dilution of Precision~(DOP) as used in the field of Global Navigation Satellite Systems (GNSS)~\cite{morton2021position}.
Both of these metrics are minimized when inter-ranging measurements come from a diverse set of directions.
For a two-dimensional robot swarm, this means that it is preferential for robots to be spread around a circle.

Several existing methods fuse inter-ranging measurements with vision for multi-robot localization~\cite{9959470, 9896952, 9655461}.
However, the existing works focus strictly on localization and measurement filtering and do not include active path planning to increase information gain from inter-ranging measurements.
One approach obtains the 3D position of a robot swarm using UWB (ultra-wideband) inter-ranging, but uses many (4+) UWBs per robot whereas our approach only needs one UWB per robot due to our advanced mesh-ranging protocol~\cite{fishberg2024multi}.
Another approach uses the Cramer-Rao lower bound in a multi-robot scenario to optimize the geometry of ranging measurements from support robots to a mission robot, but does not perform any additional sensor fusion and uses simulated ranging measurements~\cite{gipson2020swarm}.

\begin{figure}[h]
    \raggedleft
    \centering
    \includegraphics[width=0.6\linewidth]{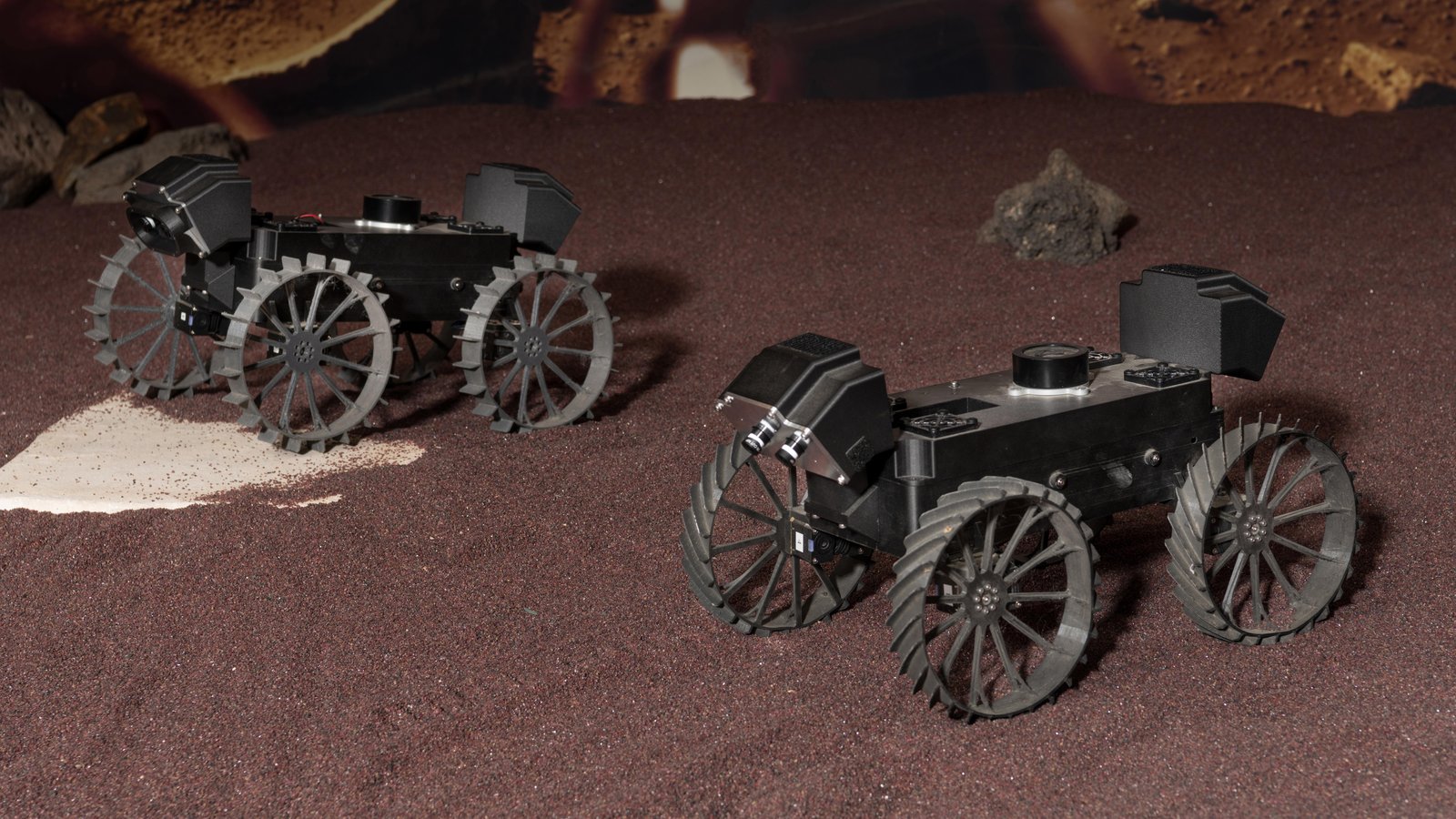}
    \caption{CADRE robot prototypes drive in formation \cite{cadre_photo}.}
    \label{fig:cadre_photo}
\end{figure}

One application where multi-robot coordination will be useful is for an upcoming National Aeronautics and Space Administration (NASA) mission called Cooperative Autonomous Distributed Robotic Exploration (CADRE) that is sending an autonomous multi-robot team to explore the surface of the Moon~\cite{Vitug2021, imnews}.
The CADRE robot swarm is made up of robots shown in Figure~\ref{fig:cadre_photo} that will collect scientific data.
The CADRE robots will be equipped with cameras to perform visual odometry, but the lunar environment is prone to visual localization errors due to its low-textured surface and shadowed regions.
To address these potential errors, the CADRE robots are equipped with UWB radios to measure inter-ranging distances.
In addition, to increase the robustness of the system against robots getting stuck or losing power, there is also a desire to use a decentralized planning algorithm that can handle robots leaving the network unexpectedly.

In this paper, we propose a novel navigation method to reduce localization error in robot swarms which leverages inter-ranging measurements from neighboring robots and plans movement paths that are optimized for swarm geometry.
To the best of our knowledge, our proposed algorithm is the first to fuse inter-ranging measurements and visual odometry for decentralized planning and estimation.
Our proposed work integrates both perception and planning strategies, which allows the multi-robot team to not only make use of inter-ranging measurements for localization, but also increases the available information from those inter-ranging measurements by adjusting the planned paths of the robots.
While this work focuses on the Cooperative Autonomous Distributed Robotic Exploration (CADRE) mission of a team of robots exploring the Moon, our navigation strategy could extend to other types of autonomous robot teams which have the capability of communicating with each other.
Other examples of autonomous teams that could leverage our approach include precision agriculture equipment, fleets of autonomous vehicles, and satellite swarms.

 Our key contributions in this paper are
 1)~we design a perception filter that uses an advanced mesh-ranging protocol to share UWB inter-ranging distance measurements simultaneously between all robots,
 2)~we develop a decentralized multi-robot coordination algorithm that minimizes localization error by optimizing planned paths for swarm geometry, and
 3)~we validate the proposed algorithm both in simulation and hardware, qualitatively observing emergent swarm formation behavior and quantitatively reducing the dilution of precision (DOP) metric and localization error.

We discuss the perception and waypoint planning portions of our proposed multi-robot coordination algorithm in Section~\ref{jpl:perception} and Section~\ref{jpl:planning} respectively.
In Section~\ref{jpl:results}, we present results from testing our multi-robot algorithm against a geometry-unaware planner both in simulation and on hardware.

\section{Perception and Localization}
\label{jpl:perception}

%In this section, we present the perception capabilities and the localization approach of each robot.
The perception capabilities of our simulated and hardware experimental robots are modeled after the lunar robots on NASA's CADRE mission~\cite{Vitug2021}.
Each robot is equipped with a stereo camera which computes visual odometry (VO).
The robots also have a sun sensor that allows them to measure the azimuth of the sun and thus compute an estimate for their absolute heading on the surface of the Moon.
Finally, each robot is outfitted with an ultra-wideband (UWB) ranging radio, which measures inter-robot distances.

In the following sections, we discuss the visual odometry, sun sensor, and UWB measurements in detail.
We further discuss the assumed communication capabilities of the robots in Section~\ref{jpl:perception:comms} and outline the extended Kalman filter used to fuse all sensor measurements in Section~\ref{jpl:perception:ekf}.

\subsection{Visual Odometry}
\label{jpl:perception:vo}

\begin{figure}[h]
    \centering
    \includegraphics[width=\linewidth]{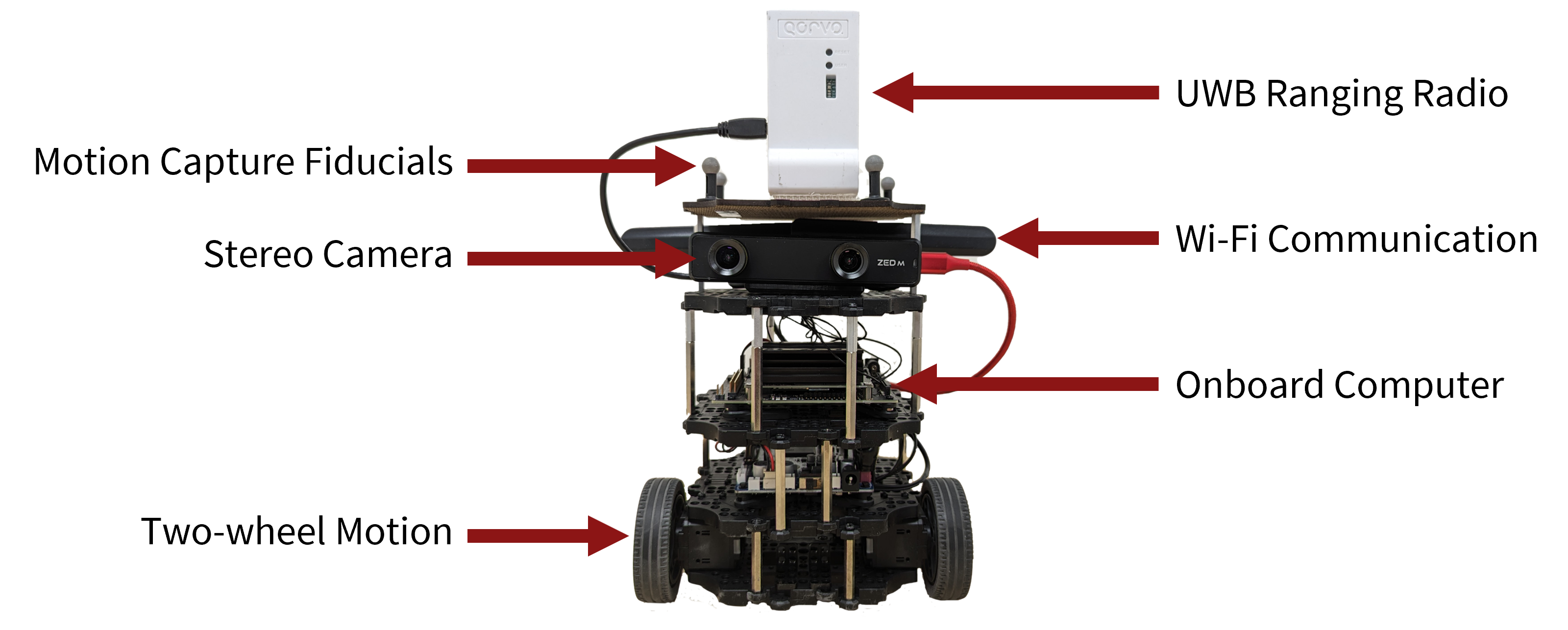}
    \caption{Robot and its components used in hardware experiments.}
    \label{fig:turtlebot_hardware}
\end{figure}

\begin{figure}[htp!]%[htp]

    \centering 

  \includegraphics[clip,width=0.7\columnwidth]{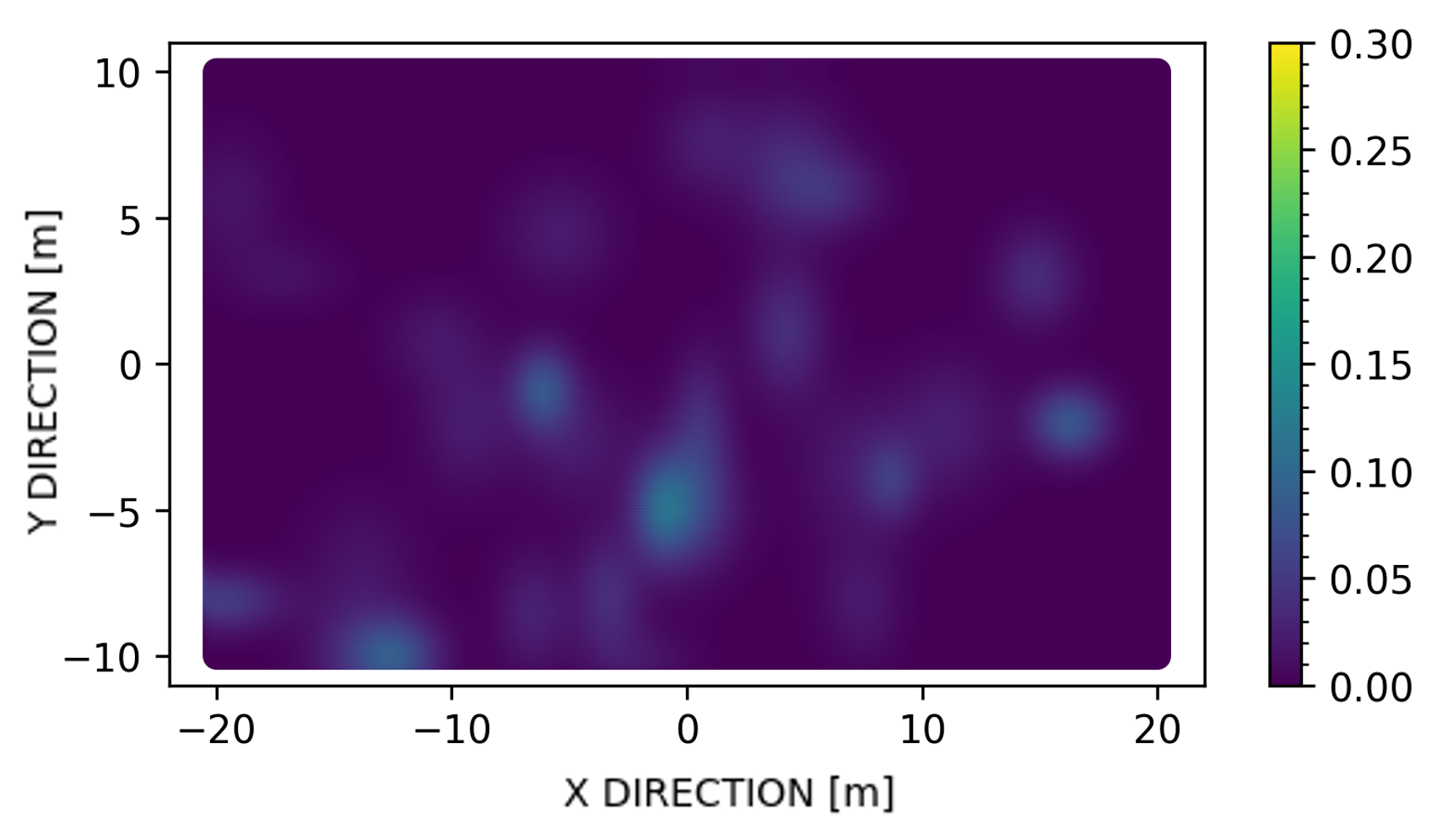}
  \medskip

  \includegraphics[clip,width=0.7\columnwidth]{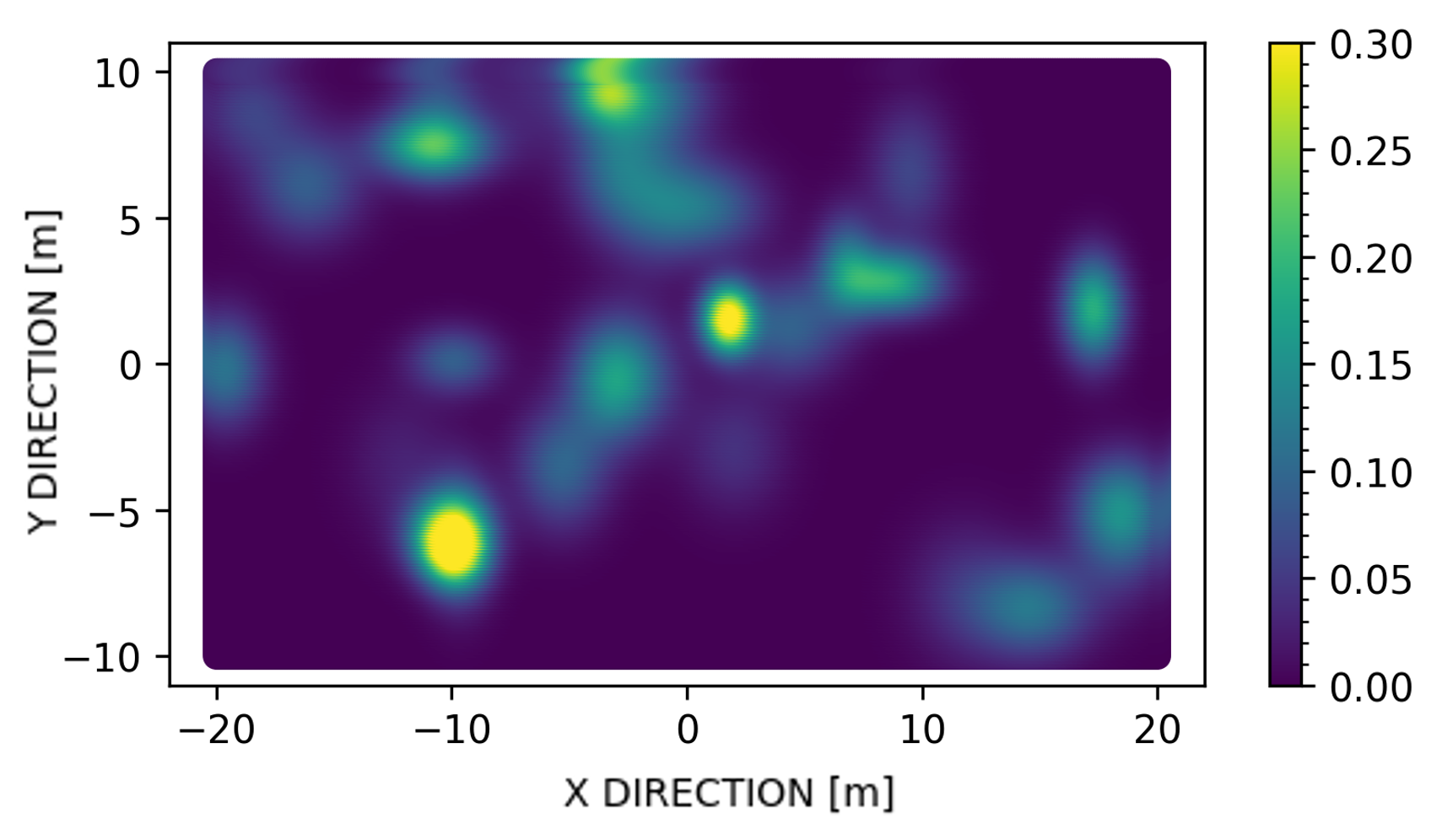}
  
  \caption{Standard deviation map of the noise added to the simulated visual odometry when the sum of the robot's linear and angular velocities is 0.2 (top) or 0.6 (bottom). Colorbar is in units of meters.}
  \label{fig:jpl:perception:vo_map}
\end{figure}

Each physical robot in our experiments as shown in Figure~\ref{fig:turtlebot_hardware} has a ZED mini stereo camera.
We use the off-the-shelf visual odometry solution available through the ZED camera's Robot Operating System (ROS) integration~\cite{github_zed}.
The visual odometry solution provided by the ZED camera outputs the position of the robot relative to its initial starting position, so we use a motion capture system to provide the visual odometry with an accurate initial position.
In simulation, we simulate visual odometry using the ground truth position of the robot and add noise that is a function of the simulated robot's velocity.
As the robot's velocity increases, the standard deviation of the noise added to the robot's position grows.
Figure~\ref{fig:jpl:perception:vo_map} shows how the standard deviation of the added noise grows with both the sum of the magnitude of the linear and angular velocities and where the robots are located within the space available for the robots to navigate.
Figure~\ref{fig:jpl:perception:vo_map} shows that in some locations the visual odometry rapidly diverges from the ground truth position while in other locations there is little added noise to the visual odometry measurements.
This feature of our simulated visual odometry attempts to mimic the phenomenon in which real-world visual odometry spatially varies in accuracy based on the available features to track as discussed in this paper's introduction.

\subsection{UWB Sensors}
\label{jpl:perception:uwb}

Typically, UWB sensors must be set up as either a static ``anchor" node or a mobile ``tag" node~\cite{stier2022}.
In that setup, the ``tag" node receives distance measurements to all ``anchor" nodes, but ``anchor" nodes do not measure the distance between themselves and other ``anchor" nodes.
Since it is desirable for all robots in the multi-robot swarm to continuously measure the distances to all other robots in the swarm, the standard UWB ``anchor" and ``tag" architecture is suboptimal.
Instead, we developed software to use custom firmware built by NASA JPL specifically for the CADRE mission that allows all robots to simultaneously measure distances between themselves~\cite{github_nasa_firmware}.
The software we created to integrate NASA JPL's UWB firmware and share distance measurements between all robots is available open source\footnote{\url{https://github.com/Stanford-NavLab/navlab_turtlebot}}.

In simulation, we compute UWB ranging measurements between the robots by using the ground truth position of each robot to compute the measured distances and publish those distances as simulated measurements.

\subsection{Sun Sensor}
\label{jpl:perception:sun}

Since both simulation and hardware demonstrations were performed indoors, we simulate the sun sensor reading.
We use either the simulation ground truth robot pose or the pose of the robot as measured by the motion capture system to obtain a ground truth yaw angle for each robot.
We then add zero mean Gaussian noise to the ground truth yaw angle and publish the measurement as the sun sensor output.

\subsection{Inter-Robot Communication}
\label{jpl:perception:comms}
In terms of communication, we assume each robot communicates with nearby robots and has its clock synchronized with its neighbors.
Each robot communicates its current position as estimated by its internal extended Kalman filter as outlined in Section~\ref{jpl:perception:ekf} and the waypoints in its latest navigation plan developed using the method discussed in Section~\ref{jpl:planning}.
Communication on hardware and in simulation happens over a Robot Operating System (ROS) network where the robots subscribe to each other's position and navigation plan topics.

\subsection{Extended Kalman Filter}
\label{jpl:perception:ekf}

Measurements from all sensors are fused using a loosely-coupled extended Kalman filter (EKF)~\cite{Kalman1960}.
%shown in Algorithm~\ref{alg:ekf}.
Similar to the parameters of the CADRE mission~\cite{Vitug2021}, we assume that the robots begin with an estimate of their absolute position that is used to initialize the extended Kalman filter.
The measurement update step of the extended Kalman filter fuses together the UWB measurements, the yaw angle from the simulated sun sensor, and the $x$ and $y$ position of the robot from the visual odometry.

The robot state, $\textbf{x}_{rx}$, estimated with the extended Kalman filter on each robot consists of its own $x$ position ($x_{rx}$), $y$ position ($y_{rx}$), and yaw angle ($\psi_{rx}$).
\begin{equation}
    \textbf{x}_{rx} = 
    \begin{bmatrix}
        x_{rx} & y_{rx} & \psi_{rx}
    \end{bmatrix}
\end{equation}

The dynamics prediction step is updated using the time since the last prediction step, $\Delta_t$, and the robot's commanded linear velocity, $v_t$, and angular velocity, $\omega_t$.

\begin{equation}
    \bar{\textbf{x}}_{rx} = 
    \textbf{x}_{rx} +
    \begin{bmatrix}
        v_t \Delta_t \cos{(\psi_{rx})} \\
        v_t \Delta_t \sin{(\psi_{rx})} \\
        \omega_t \Delta_t \\
    \end{bmatrix}
\end{equation}

We assume that the measurement vector for the EKF consists of the concatenated UWB distance measurements ($UWB_{n}$),
the visual odometry $x$ position ($x_{vo}$) and $y$ position ($y_{vo}$), and the sun sensor angle output ($\psi_{sun}$).
The measurement vector for the case of $n$ neighboring robots is shown in Equation~\ref{eq:jpl:perception:zt}.
\begin{equation}
    \label{eq:jpl:perception:zt}
    \textbf{z}_{t} = 
    \begin{bmatrix}
        UWB_{1} & \cdots & UWB_{n} & x_{vo} & y_{vo} & \psi_{sun}
    \end{bmatrix}
\end{equation}
The measurement Jacobian from the EKF update step is shown in Equation~\ref{eq:jpl:perception:ht}.
$x_i$ and $y_i$ refers to the $x$ and $y$ position of the $i$th neighbor.
\begin{equation}
    \label{eq:jpl:perception:ht}
    H_t = 
    \begin{bmatrix}
        \frac{x_{rx} - x_1}{r_1} & \frac{y_{rx} - y_1}{r_1} & 0 \\
        \vdots & \vdots & \vdots \\
        \frac{x_{rx} - x_n}{r_n} & \frac{y_{rx} - y_n}{r_n} & 0 \\       
        1 & 0 & 0 \\
        0 & 1 & 0 \\
        0 & 0 & 1 \\
    \end{bmatrix}
\end{equation}

\begin{equation}
    r_i = \sqrt{(x_{rx} - x_{i})^2 + (y_{rx} - y_{i})^2}
\end{equation}

\section{Waypoint Planning Approach}
\label{jpl:planning}

In this section, we introduce the waypoint planning approach of our multi-robot coordination algorithm.
The algorithm uses a planner that selects a waypoint to navigate towards, passes that waypoint to a local planner, and then replans to select a new waypoint during the execution of the current plan.
To choose the next waypoint, each robot samples points in its immediate vicinity and computes a cost for each sampled point.
The total cost for each potential waypoint consists of a combination of a goal cost, collision gost, and DOP cost, which we detail in the sections that follow. 
The planning approach is decentralized meaning that each robot determines its next waypoint by itself using communication only to exchange the estimated position and planned paths of the other robots.

\subsection{Goal Cost}
\label{jpl:planning:goal}

The goal cost is defined as the Euclidean distance between the potential waypoint and the robot's final goal location. As a final step, the goal cost is normalized by the maximum goal cost value so that the goal cost always ranges between zero and one.
\begin{equation}
    \label{jpl:goal_cost}
    c_{goal} = \norm{\textbf{x}_{rx} - \textbf{x}_{goal}}
\end{equation}

\subsection{Collision Cost}
\label{jpl:planning:collision}

The collision cost makes use of the time-synchronized future planned paths of the neighboring robots.
The collision cost is defined in Equation~\ref{jpl:collision_cost} where $d_{min}$ is equal to the minimum distance to any neighboring robots in future timesteps, $d_{avoid}$ adjusts the desired avoidance radius between robots,  and $\epsilon$ is an arbitrary small value to prevent division by zero.
\begin{equation}
    \label{jpl:collision_cost}
    c_{collision} = \frac{d_{avoid}}{d_{min} + \epsilon}
\end{equation}

The collision cost increases as the minimum distance approaches zero.
For our implementation we use $d_{avoid} = 0.4$ since we desire the robots to maintain a separation of approximately 0.4 meters away from each other.
This cost function discourages robots from approaching near other robots and has a negligible small value as the robots are spaced further apart.

\subsection{DOP Cost}
\label{jpl:planning:dop}

The DOP cost is based off of the idea of dilution of precision~\cite{morton2021position}, and prioritizes creating geometries that increase the information gain of inter-ranging measurements.
The cost expression, given in Equation~\ref{jpl:eq:dop_cost}, is computed from $A$, the matrix of normalized line-of-sight vectors to each of the $n$ neighboring robots.
\begin{equation}
    \label{jpl:eq:dop_cost}
    c_{DOP} = \sum_{i \in \mathcal{N}} \sqrt{trace\left(\invsub{A^\top A}\right)}
\end{equation}
\begin{equation}
    A = 
    \begin{bmatrix}
        \frac{x_1 - x_{rx}}{\sqrt{(x_1 - x_{rx})^2 + (y_1 - y_{rx})^2}} & 
        \frac{y_1 - y_{rx}}{\sqrt{(x_1 - x)^2 + (y_1 - y_{rx})^2}} \\        
        \vdots & \vdots \\
        \frac{x_n - x_{rx}}{\sqrt{(x_n - x_{rx})^2 + (y_n - y_{rx})^2}} & 
        \frac{y_n - y_{rx}}{\sqrt{(x_n - x_{rx})^2 + (y_n - y_{rx})^2}} \\
    \end{bmatrix}
\end{equation}

The DOP cost from Equation~\ref{jpl:eq:dop_cost} is calculated at each time-synchronized step of the known future plans for the robots.
At each timestep, the DOP cost is computed from the perspective of each robot in the neighbor set ($\mathcal{N}$) including itself and summed together.
The final DOP cost is taken as the maximum value across all timesteps.
As a final step, the DOP cost is normalized by the median DOP value from the set of potential waypoints to ensure that at least half of the waypoints have a cost between zero and one.
Normalizing allows the goal cost and DOP cost to have similar magnitudes.
\begin{figure}[h]
    \raggedleft
    \centering
    \includegraphics[width=0.8\linewidth]{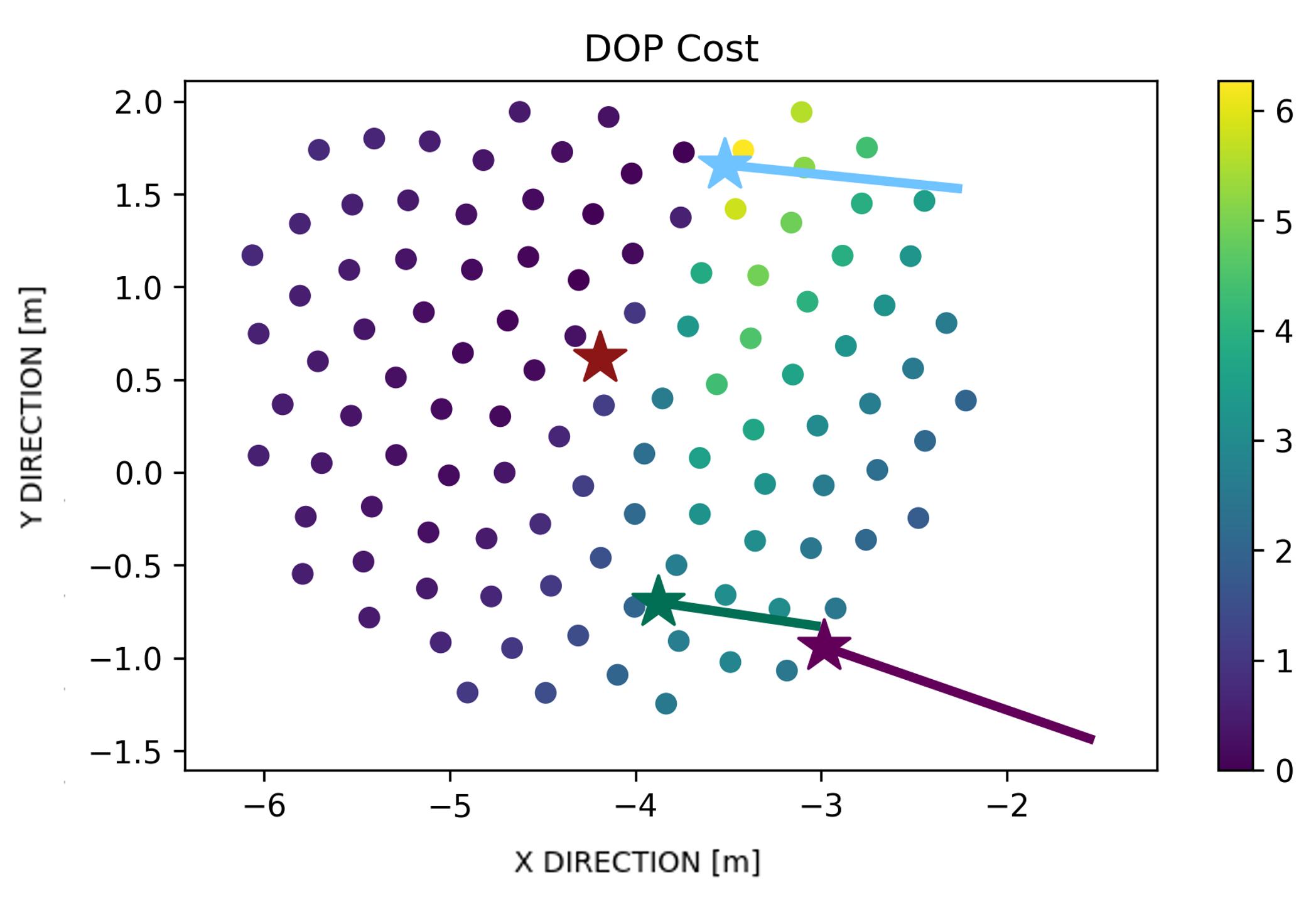}
    \caption{DOP costs for potential waypoints for middle robot whose position is marked with a star. The neighboring robots' current positions are marked with a star and their communicated future paths are drawn with lines.}
    \label{fig:jpl:planning:dop:example1}
\end{figure}
Figure~\ref{fig:jpl:planning:dop:example1} shows an example of the DOP cost for potential waypoints for the robot in the center.
Potential waypoints which are on the right side and create more of a linear geometry have a much higher DOP cost than the potential waypoints on the left side that would create a more circular geometry among the four robots.

\subsection{Total Cost}
\label{jpl:planning:total}

The total cost for each potential waypoint consists of a linear combination of all three aforementioned costs and shown in Equation~\ref{jpl:total_cost}.
\begin{equation}
    \label{jpl:total_cost}
    c_{total} = \alpha c_{DOP} + \beta c_{goal} + c_{collision}
\end{equation}
To weigh progress toward goals versus improving the swarm geometry to minimize localization error, we use the weighting variables $\alpha$ and $\beta$ that are based on the maximum uncertainty from all robots in the neighboring set $\mathcal{N}$.
The maximum uncertainty is calculated as a combination of the $x$-direction and $y$-direction elements of the covariance matrix from the EKF discussed in Section~\ref{jpl:perception:ekf}.
\begin{equation}
    \label{jpl:eq:alpha}
    \alpha = \frac{1}{1 + e^{5-10 \Sigma_{max}}}
\end{equation}
\begin{equation}
    \Sigma_{max} = \max_{\forall i \in \mathcal{N}} \sqrt{\Sigma_{xx} + \Sigma_{yy}}
\end{equation}
\begin{equation}
    \label{jpl:eq:beta}
    \beta = 1 - \alpha
\end{equation}
Equation~\ref{jpl:eq:alpha} and Equation~\ref{jpl:eq:beta} are defined such that as the maximum uncertainty of the robot swarm approaches one meter, the robot swarm prioritizes minimizing the DOP cost to allow for lowering the localization uncertainty of the system.
\begin{figure}[h]
    \raggedleft
    \centering
    \includegraphics[width=0.8\linewidth]{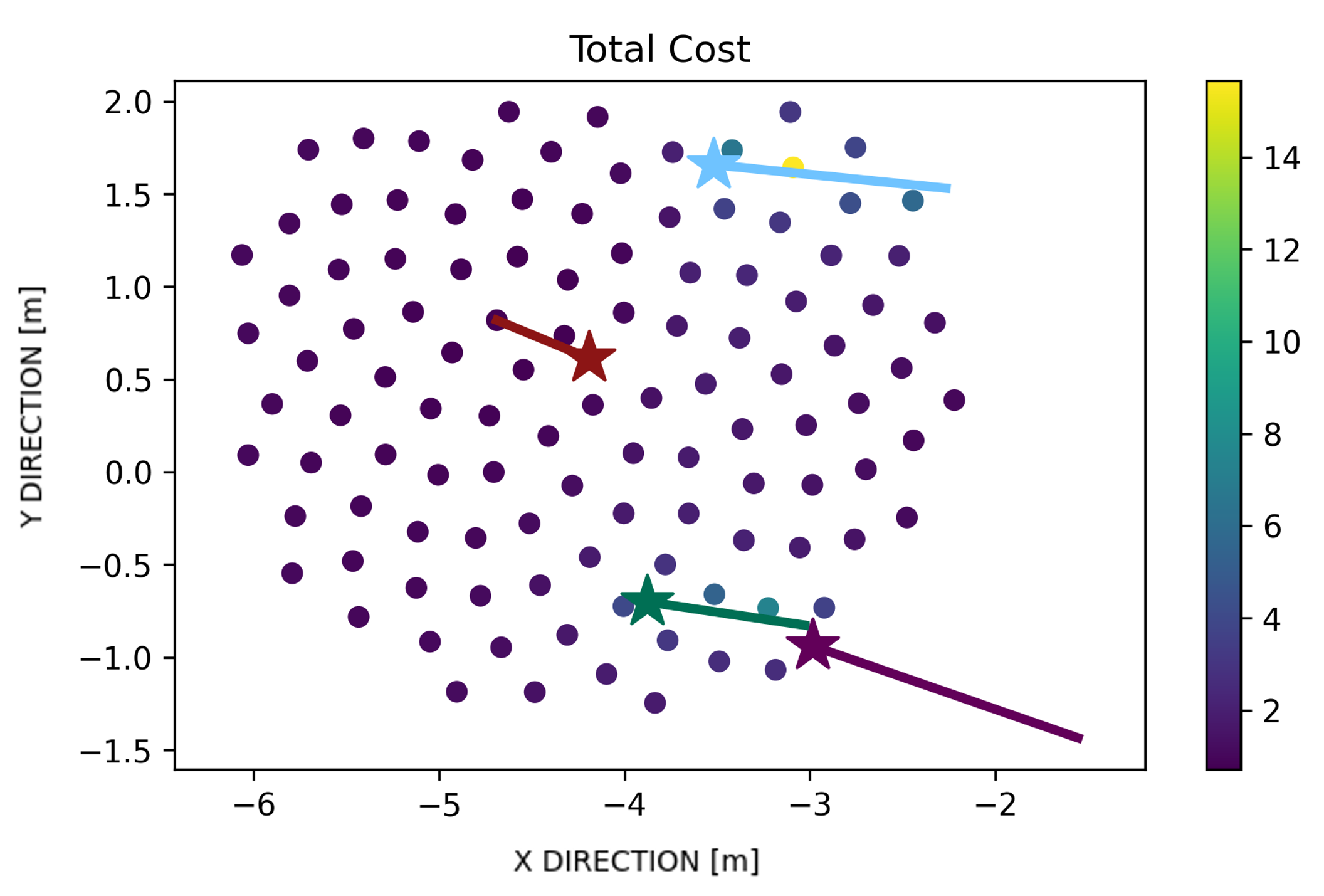}
    \caption{Total costs for potential waypoints for middle robot whose position is marked with a star. The neighboring robots' current positions are marked with a star and their communicated future paths are drawn with lines.}
    \label{fig:jpl:planning:total:example1}
\end{figure}
Once all costs are combined, the robot chooses to navigate towards the potential waypoint with the lowest cost and communicates its plan to the neighboring robots.
Figure~\ref{fig:jpl:planning:total:example1} shows the total waypoint cost for the middle robot.
In this scenario, the value of $\alpha$ was large enough that the robot prioritized improving the geometry of the system over heading directly towards its goal on the ride side of the graph.

\section{Results}
\label{jpl:results}

To validate the proposed multi-robot coordination algorithm, we compare against a naive algorithm that routes robots directly to their respective goals and does not optimize for geometry.
For each test, four robots started in initial positions in a straight line and navigated across a room to final goal locations.
The code used for both simulation and hardware results is available open source~\cite{github_navlab_turtlebot}.

\subsection{Simulation Results}
\label{jpl:results:sim}

\begin{figure}[h]
    \raggedleft
    \centering
    \includegraphics[width=\linewidth]{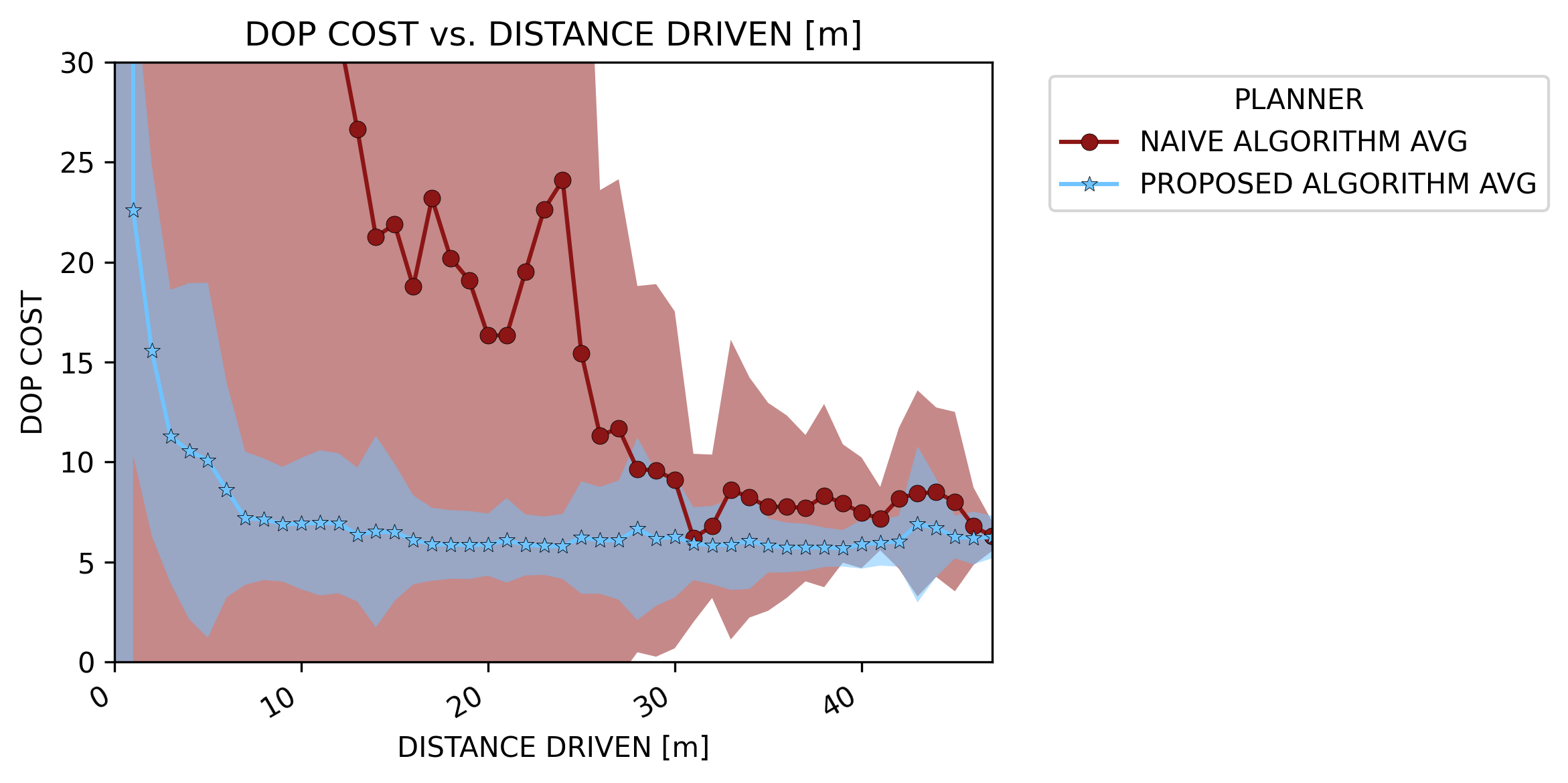}
    \caption{DOP cost over distance in simulation. One standard deviation on either side of the mean is shaded on the plot.}
    \label{fig:sim_dop}
\end{figure}

We use Gazebo for simulation, with high-fidelity robot models matching those in hardware~\cite{gazebo}.
Figure~\ref{fig:sim_dop} shows a quantitative metric for the robots' geometry over time.
The graph plots the DOP cost from Equation~\ref{jpl:eq:dop_cost} averaged across all four robots and 30 experimental runs as a function of the total number of meters driven by the robot swarm.
%The mean of all four robots is plotted and one standard deviation on either side of the mean is shaded.
The figure shows that as the robots traversed towards their goals, the naive algorithm had a much higher DOP cost illustrating the fact that when the robots use the proposed multi-robot coordination algorithm, the robots are able to organize into geometries that are conducive to higher information gain.

\begin{figure}[h]
    \raggedleft
    \centering
    \includegraphics[width=\linewidth]{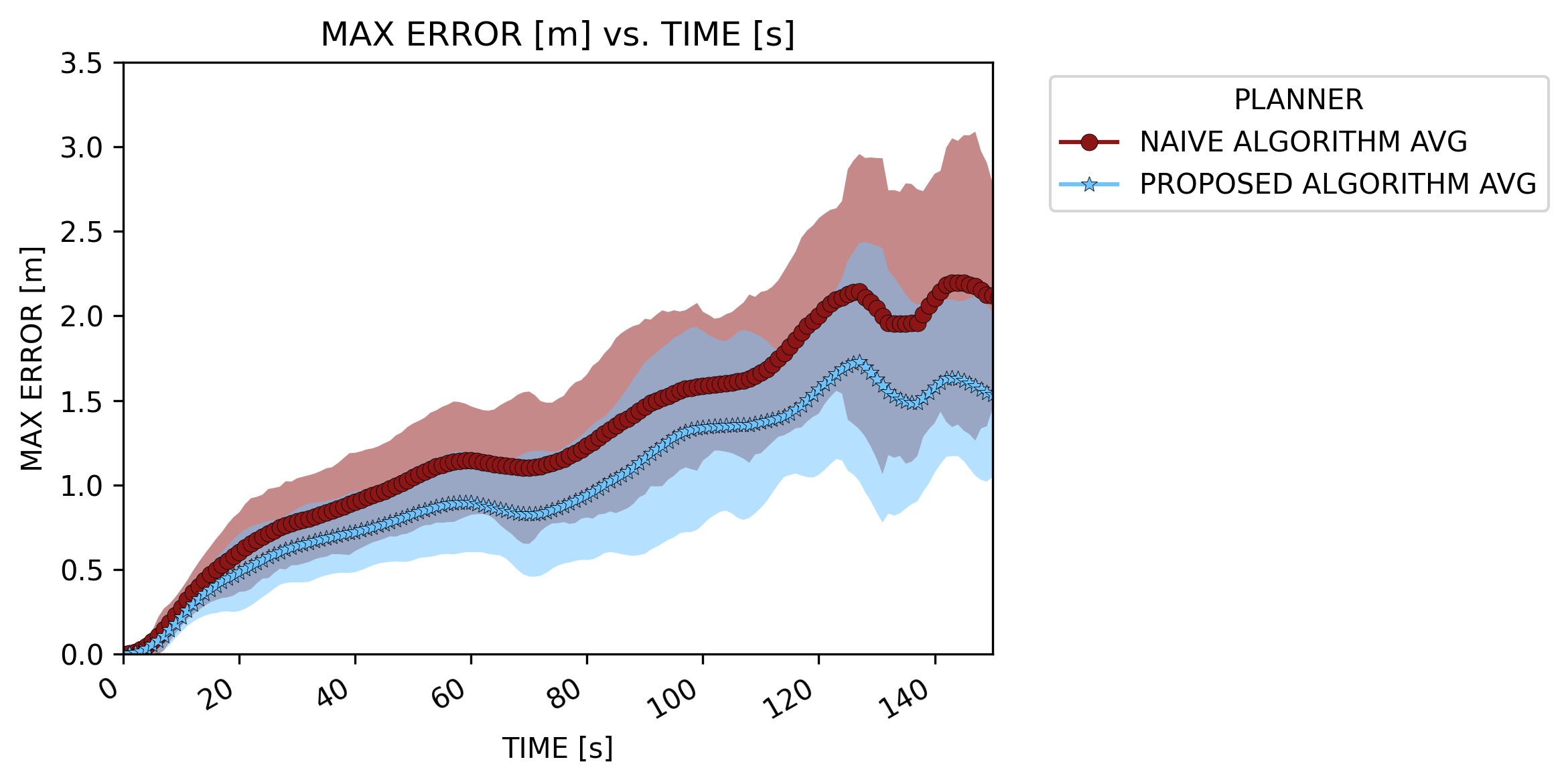}
    \caption{Maximum position error in simulation. One standard deviation on either side of the mean is shaded on the plot.}
    \label{fig:sim_clir_error_max}
\end{figure}

To compute a localization accuracy metric, we compared the estimated position by either the naive algorithm or our proposed algorithm with the ground truth position as output by the Gazebo simulator.
Simulation results revealed that using the proposed multi-robot coordination algorithm decreased both the mean and maximum position error as averaged across each of the four robots.
%Figure~\ref{fig:sim_clir_error_mean} and 
Figure~\ref{fig:sim_clir_error_max} shows the maximum position error respectively averaged across each of the four robots and averaged across 30 simulation experiments for each type of algorithm.

\subsection{Hardware Experiment Results}
\label{jpl:results:hard}

For hardware experiments, we used physical twins of the CADRE robots with similar perception capabilities, shown in Figure~\ref{fig:turtlebot_hardware}.
The robots' sensors include a DWM1001-DEV UWB ranging radio, Jetson Nano onboard computer, and a ZED mini stereo camera.
We perform our experiments in the Stanford Flight Room, with ground truth available through an OptiTrack motion capture system.

\begin{figure}[h]%[htp]

\centering 

  \includegraphics[clip,width=0.7\columnwidth]{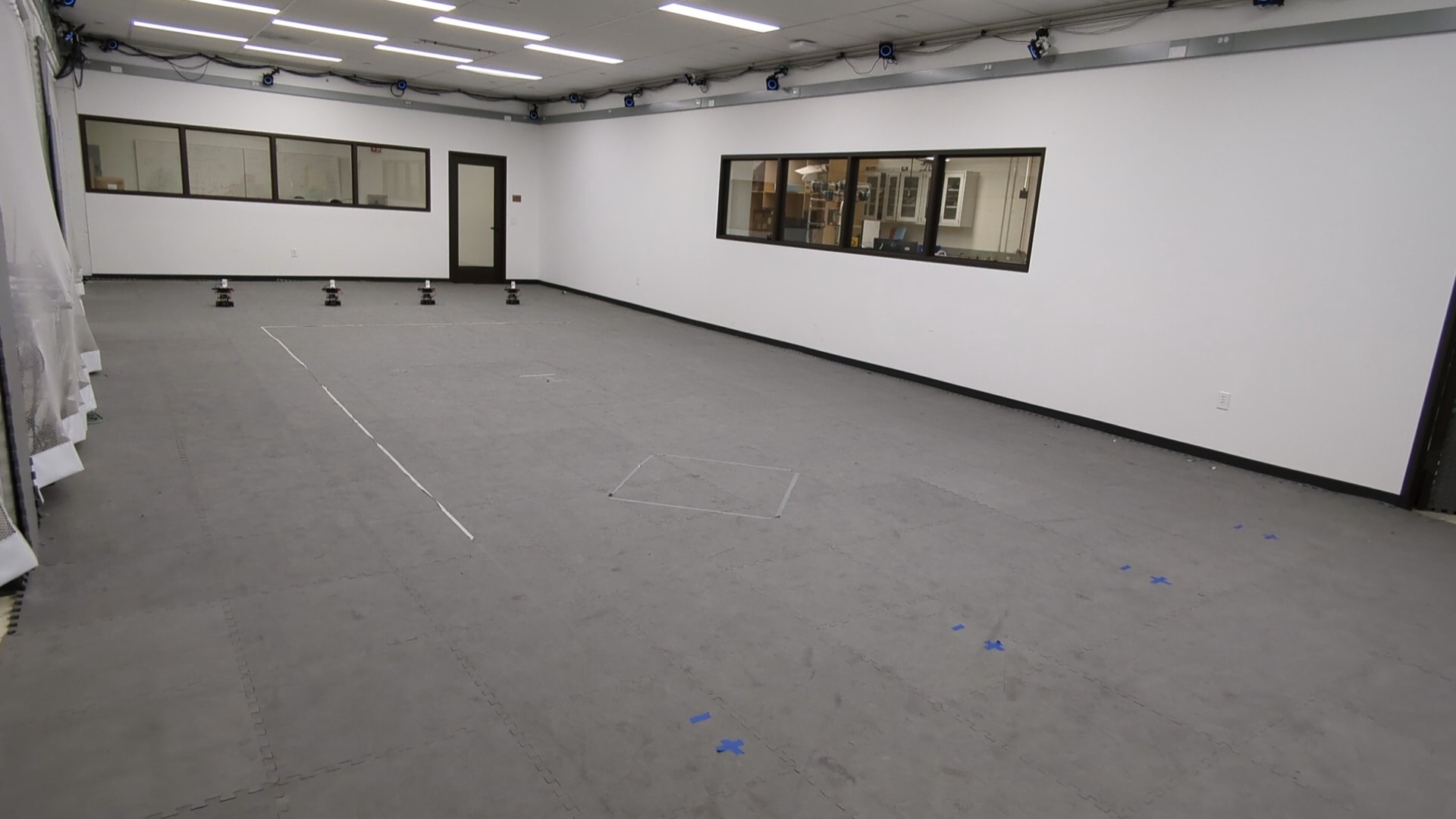}%
  \medskip

  \includegraphics[clip,width=0.7\columnwidth]{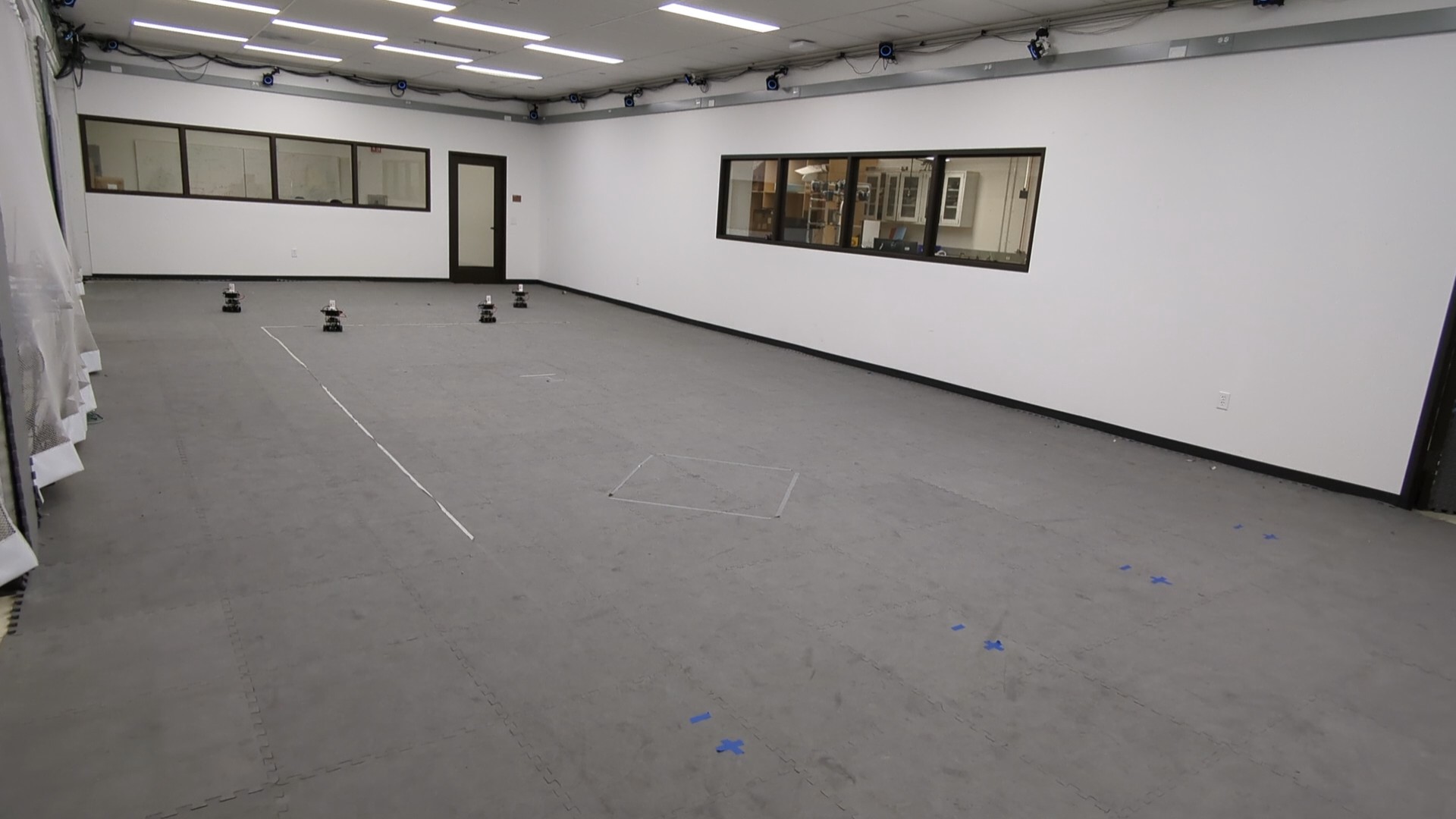}%
  \medskip
  
  \includegraphics[clip,width=0.7\columnwidth]{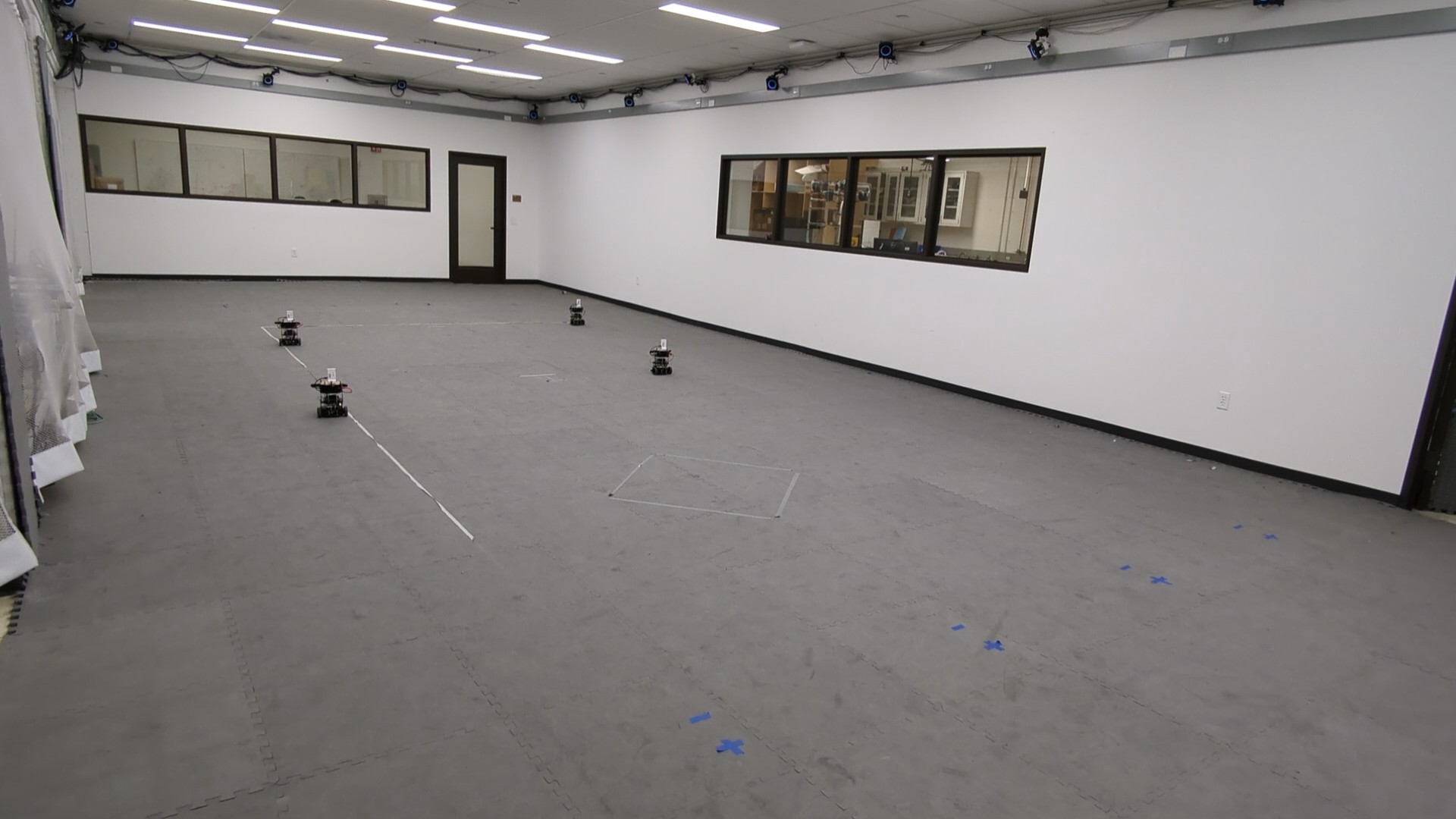}%
  % \medskip
  
  % \includegraphics[clip,width=0.6\columnwidth]{figures/hardware_geo_4.jpg}%

\caption{Robots using the proposed multi-robot coordination algorithm form optimal square geometry for minimizing localization error while traversing towards their respective goal locations.}
\label{fig:jpl:hardware_run_clir}

\end{figure}

With the naive planner, the robots do not optimize for geometry, but rather path directly towards their respective goals.
On the other hand, when using the proposed multi-robot coordination algorithm, we found that the robots moved into a square formation to traverse across the room and towards their goals rather than remaining in a straight line.
A square formation is the geometry that minimizes the DOP cost from Equation~\ref{jpl:eq:dop_cost} and arose as an emergent property of the proposed algorithm's cost function rather than an explicit command of such a formation. Figure~\ref{fig:jpl:hardware_run_clir} shows an example hardware experiment with the robots starting in a straight line and then moving into a square formation to traverse across the room with improved geometry.

\begin{figure}[h]
    \raggedleft
    \centering
    \includegraphics[width=\linewidth]{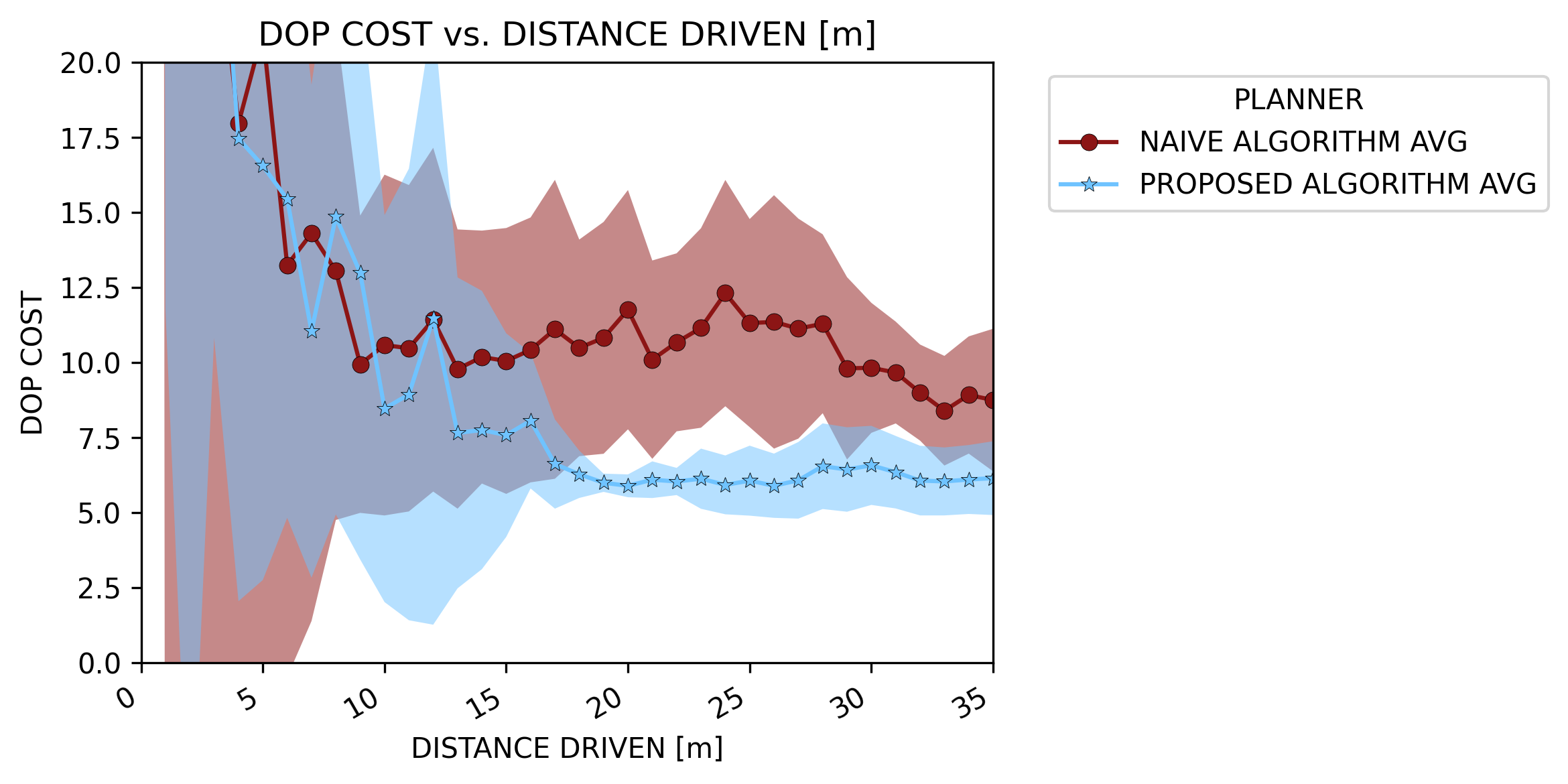}
    \caption{DOP cost over distance on hardware. One standard deviation on either side of the mean is shaded on the plot.}
    \label{fig:hardware_dop}
\end{figure}

The enhanced geometry using the proposed multi-robot coordination algorithm was again validated on hardware using the DOP cost from Equation~\ref{jpl:eq:dop_cost} averaged across all four robots and the six hardware experiments for each of the two types of planning algorithms.
We show in Figure~\ref{fig:hardware_dop} that the proposed algorithm is able to maintain lower DOP cost and thus provide inter-ranging measurements with more information gain than the naive algorithm.
In hardware, there was a negligible difference in localization error between the naive and proposed planners due to inherent motion and sensing uncertainties from the hardware, which will be addressed in future work.

\section{Conclusion}
\label{jpl:summary}

In this paper, we proposed a novel active localization strategy that not only uses inter-ranging measurements to benefit the robots' localization, but also actively plan paths that increase the information gain of those inter-ranging measurements through enhanced geometry.
The proposed method is decentralized, in that robots may leave or join the network without interruption to the perception or planning algorithms.
We showed in both simulation and on hardware that the proposed multi-robot coordination algorithm minimizes the dilution of precision metric and reduces localization error when compared with a naive planner that does not make use of inter-ranging measurements.
The proposed planner could be further scaled beyond the CADRE mission to other types of multi-robot swarms capable of communication and inter-ranging.

\section*{Acknowledgment}
 
This work was supported by the National Science Foundation (NSF) [award number 2006162] and by NASA Jet Propulsion Laboratory’s (JPL’s) Strategic University Research Partnership program.

\bibliographystyle{IEEEtran}
\bibliography{references}

\end{document}